\documentclass[letterpaper, 10 pt, preprint, conference]{ieeeconf}  

\IEEEoverridecommandlockouts                              

\usepackage{comment}
\usepackage{gensymb}
\usepackage[hidelinks]{hyperref}
\usepackage{graphicx}       
\usepackage{caption}        
\usepackage{float}          
\usepackage{amsmath}
\usepackage{amsfonts}
\usepackage{tabularx}
\hypersetup{bookmarks=false}

\title{\LARGE \bf
Finding an Initial Probe Pose in Teleoperated Robotic Echocardiography via 2D LiDAR-Based 3D Reconstruction
}

\usepackage{fancyhdr}

\author{
	Mariadas Capsran Roshan$^{1}$*, Edgar M Hidalgo$^{1}$, Mats Isakssson$^{1}$, Michelle Dunn$^{1}$, Jagannatha Charjee Pyaraka$^{1}$%
	\thanks{*Corresponding author\newline $^{1}$School of Engineering, Swinburne University of Technology, Victoria, Australia. Emails: \tt\small mroshan@swin.edu.au, ehidalgoflorez@swin.edu.au, misaksson@swin.edu.au, jdunn@swin.edu.au, jagannathacharjeepya@swin.edu.au}
}

\begin{document}

\maketitle

\maketitle
\fancypagestyle{firstpage}{
	\fancyhf{}
	\renewcommand{\headrulewidth}{0pt}
	\fancyhead[C]{\small This work has been submitted to the IEEE for possible publication. 
		Copyright may be transferred without notice, after which this version 
		may no longer be accessible.}
}
\thispagestyle{firstpage}

\begin{abstract}

Echocardiography is a key imaging modality for cardiac assessment but remains highly operator-dependent, and access to trained sonographers is limited in underserved settings. Teleoperated robotic echocardiography has been proposed as a solution, however, clinical studies report longer examination times than manual procedures, increasing diagnostic delays and operator workload. Automating non-expert tasks, such as automatically moving the probe to an ideal starting pose, offers a pathway to reduce this burden. Prior vision- and depth-based approaches to estimate an initial probe pose are sensitive to lighting, texture, and anatomical variability. We propose a robot-mounted 2D LiDAR-based approach that reconstructs the chest surface in 3D and estimates the initial probe pose automatically. To the best of our knowledge, this is the first demonstration of robot-mounted 2D LiDAR used for 3D reconstruction of a human body surface. Through plane-based extrinsic calibration, the transformation between the LiDAR and robot base frames was estimated with an overall root mean square (RMS) residual of~1.82 mm and rotational uncertainty below 0.2\degree. The chest front surface, reconstructed from two linear LiDAR sweeps, was aligned with non-rigid templates to identify an initial probe pose. Mannequin-based study assessing reconstruction accuracy showed mean surface errors of 2.78$\pm$0.21 mm. Human trials (N=5) evaluating the proposed approach found probe initial points typically 20--30 mm from the clinically defined initial point, while the variation across repeated trials on the same subject was less than 4 mm.

\end{abstract}

\section{INTRODUCTION}

Echocardiography is the application of ultrasound to visualize the heart, providing real-time assessment of cardiac anatomy and function. It is a primary diagnostic tool for detecting heart failure, evaluating valvular disorders, identifying congenital abnormalities, and guiding decision-making in emergency and critical-care contexts~\cite{EchocardiographyHeartFailure2012,lancellottiUseEchocardiographyAcute2015}. The accuracy of echocardiography is highly operator-dependent, and access to trained sonographers is limited in regional and underserved areas~\cite{roshanRoboticUltrasonographyAutonomous2022b}. Delays in cardiac assessment worsen outcomes, with one study~\cite{alrawashdehImpactEmergencyMedical2021} reporting that in acute heart attack patients each additional 30-minute emergency medical transport delay increased the 30-day mortality risk by 20\%.

Teleoperated robotic echocardiography~\cite{hidalgoCurrentApplicationsRobotAssisted2023b} has been proposed as a promising solution to overcome this limitation. Such a solution includes a robot equipped with an ultrasound probe is manipulated remotely by an echocardiographer, with real-time feedback enabling accurate image acquisition across distance. A clinical study~\cite{bomanRobotAssistedRemoteEchocardiographic2014a} demonstrated that this approach can reduce diagnostic delays by nearly three months compared to conventional referral pathways that often require patients to travel to specialized facilities. However, as summarized in~\autoref{tab:teleop_vs_manual}, multiple clinical studies report longer durations for teleoperated robot echocardiography compared to manual examinations.

\begin{table}[b]
	\centering
	\caption{Comparison of examination durations between teleoperated and manual echocardiography.}
	\label{tab:teleop_vs_manual}
	\begin{tabular}{p{2.1cm} p{1.8cm} p{2.0cm} p{1.0cm}}
		\hline
		\textbf{Study} & \textbf{Teleoperated Duration} & \textbf{Manual \newline Duration} & \textbf{Relative Increase} \\
		\hline
		Liu et al.~\cite{liuInitialExperience5G2024}   & 24 min 36 s          & 16 min 15 s         & $\sim$1.5$\times$ \\
		Wright et al.~\cite{wrightAssessingViabilityPatient2025} & $32.7 \pm 8.9$ min   & $5.9 \pm 0.92$ min  & $\sim$5.5$\times$ \\
		Solvin et al.~\cite{solvinFeasibilityTeleoperatedRobotic2023} & 26.4 min             & Standard bedside    & $\sim$1.9$\times$ \\
		\hline
	\end{tabular}
\end{table}

Previous research~\cite{hidalgoEvaluatingImpactsNetwork2025a,jiangRoboticUltrasoundImaging2023a} has proposed that not all stages of echocardiography require expert input and that repetitive preparatory steps can be automated to reduce examination time. One such step is the initial placement of the probe on the chest above the heart, which currently requires manual maneuvering from a neutral position. This process begins with localizing the cardiac region of interest (ROI) on the chest surface, followed by identifying a suitable position and orientation of the probe at that ROI, and finally moving the probe into this configuration. Automating this sequence can reduce procedure time and operator workload. 

Several studies have investigated vision-based approaches to automate initial probe pose estimation within teleoperated and other robot-assisted ultrasound imaging systems, spanning applications in both echocardiography and other anatomical targets. Mustafa et al.~\cite{binmustafaHumanAbdomenRecognition2013} targeted the epigastric region by processing a frontal RGB photograph of the torso, automatically detecting anatomical landmarks such as the umbilicus and nipples, and then applying anthropometric distance ratios to triangulate  an approximate target position for probe placement. In contrast, Ning et al.~\cite{ningAutonomicRoboticUltrasound2021} avoided landmarks through a reinforcement learning framework. Their approach used a state-representation network to embed ultrasound and contact-force context into the RGB frame, while a pre-trained policy consumed these cues to guide the probe toward the ROI. Both approaches highlight how RGB-only cues can initialize probe placement, though they remain sensitive to lighting and appearance variations~\cite{biMachineLearningRobotic2024}. Additionally, these approaches have not been demonstrated for cardiac imaging, where anatomical variability and motion add further challenges~\cite{tangAutonomousUltrasoundScanning2024}.

To address the limitations of RGB-based approaches, particularly their dependence on visual texture, some works employ depth sensors to capture geometric surface cues. Graumann et al.~\cite{graumannRoboticUltrasoundTrajectory2016} and Kaminski et al.~\cite{kaminskiFeasibilityRobotAssistedUltrasound2020} registered pre-segmented anatomical models (e.g., cardiac or thyroid regions) onto live patient point clouds acquired with RGB-D cameras, using geometry-based alignment techniques to estimate the ROI and probe pose. In contrast, Tan et al.~\cite{tanFlexibleFullyAutonomous2023a} fused multi-view structured-light point clouds, used keypoints to segment the ROI, and estimated the probe\rq s initial position from surface points with its orientation aligned to local surface normals. While these approaches improve robustness over RGB-only approaches, they remain constrained by depth sensing, which is vulnerable to ambient light~\cite{seewaldAnalyzingMutualInterference2019}, limited range~\cite{seoSensingRangeExtension2022}, and the need for multi-angle views or wide baselines~\cite{imamOptimisingBaselineDistance2025} that are impractical in often space-limited teleoperated settings.

To overcome the limitations of vision-based approaches, we propose a two-dimensional light detection and ranging (2D LiDAR)-based initial probe pose estimation approach. A robot-mounted 2D LiDAR performs two linear sweeps across the patient\rq s chest front surface, which, following plane-based calibration, yield a reconstructed 3D point cloud. This point cloud is preprocessed to remove noise and non-relevant structures, after which non-rigid template matching identifies the initial probe position. Finally, local plane fitting at that point estimates the surface normal to derive the probe orientation for placement. 

The LiDAR-based approach directly addresses several limitations of vision-based approaches. LiDAR measurements are based on time-of-flight ranging, making them invariant to ambient lighting~\cite{padmanabhanModelingAnalysisDirect2019}. This avoids degradation from shadows, clothing textures, or background clutter that commonly affect camera-based methods~\cite{whyteApplicationLidarTechniques2015}. Moreover, LiDAR sensing encodes spatial geometry through range measurements without capturing photometric or textural information, thereby inherently supporting privacy-preserving patient monitoring~\cite{wuLidarbasedComputerVision2023}. Additionally, robot-mounted LiDAR enables passive multi-view acquisition through platform motion, eliminating the need for wide camera baselines or fixed external sensor placements. To the best of our knowledge, this work presents the first demonstration of reconstructing a human body surface in 3D using a robot-mounted 2D LiDAR scanner.

Although 3D LiDAR systems can capture volumetric data in a single scan, even compact models like the Ouster OS0 Gen2 weigh between 445 g and 930 g~\cite{OusterOS2HighPrecision2025}, with high-resolution units exceeding 8 kg. In contrast, 2D LiDARs such as the RPLIDAR C1 weigh just 110 g~\cite{RPLIDARC1Fusion2025}, minimizing payload and inertia on robotic arms in clinical settings. Angular resolution is also finer in 2D LiDARs (0.25\degree-1.0\degree)~\cite{mazzariHowSelectRight2025} compared to 3D LiDARs' sparser vertical resolution (0.4\degree-2.0\degree)~\cite{zhouS4SLAMRealtime3D2021}.

In this work, we present the first use of robot-mounted 2D LiDAR for probe pose initialization in tele-echocardiography. Our contributions are: (1) the first demonstration of full 3D human body surface reconstruction using robot-mounted 2D LiDAR; (2) a plane-based extrinsic calibration method, achieving an overall root mean square (RMS) residual of ~1.8 mm and rotational uncertainty below 0.2\degree~for robust robot--LiDAR alignment; (3) a non-rigid template-matching framework to estimate the initial probe position and orientation; and (4) two quantitative evaluations: one assessing the accuracy of the 3D chest front surface reconstruction against high-resolution scans of male and female mannequins, and another on human participants to evaluate the accuracy and repeatability of the complete approach.

\begin{figure}[b!]
	\centering
	\includegraphics[width=0.48\textwidth, trim={0cm 0.1cm 0cm 0cm}, clip]{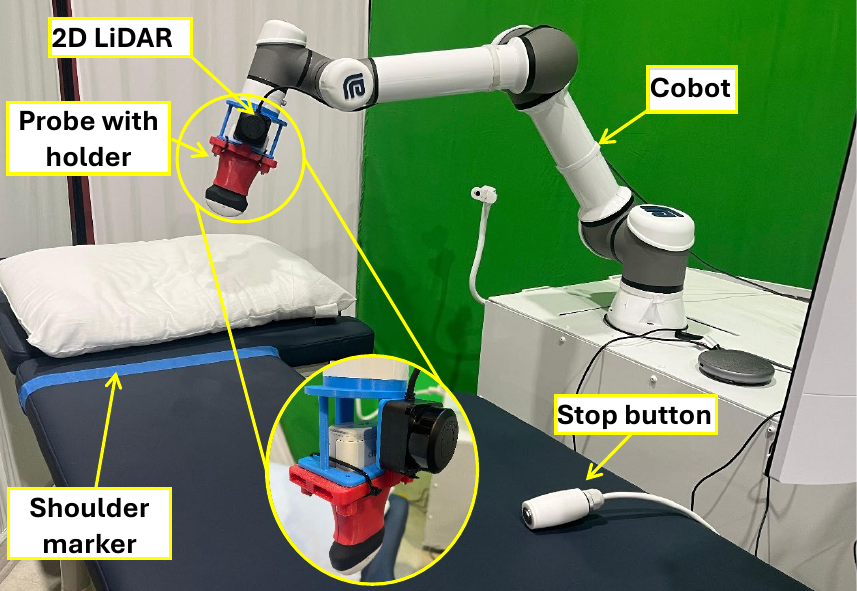}
	\captionsetup{width=\linewidth}
	\caption{System setup with UR5e cobot, probe holder with Clarius ultrasound probe, integrated 2D RPLIDAR C1 and emergency stop button. The bed includes a blue shoulder marker defining the safety plane for restricted robot motion.}
	\label{fig:fig3}
\end{figure}

\section{Methods}

\subsection{System Overview}

While the proposed LiDAR-based initial probe pose estimation approach is system-agnostic, this study uses the configuration in~\autoref{fig:fig3} for implementation and evaluation. We focus on the robot-side components, as the work targets automation of the initial probe positioning step prior to teleoperation. The core platform is a UR5e collaborative robot (cobot), chosen for its compliance and built-in safety features. A Clarius C3 HD3 wireless ultrasound probe was mounted to emulate a realistic clinical setup and to verify that the probe geometry did not interfere with the proposed approach. The probe holder from~\cite{roshanLowcostProbeChanger2022b}, originally designed for the ultrasound probe, was modified to also accommodate a 2D LiDAR sensor on the robot mount. For this study, we used a Slamtec RPLIDAR C1, a lightweight 360\degree~LiDAR with 0.72\degree~angular resolution, chosen to demonstrate that the approach can be realized with low-cost hardware rather than requiring a high-end sensor. The control architecture and data flow were implemented in Robot Operating System (ROS)~1. Safety measures include constraining the robot's tool center point (TCP) speed and joint limits in software, defining a virtual safety plane to prevent motion toward the participant's head, and providing an emergency stop button for immediate interruption in case of discomfort.

\subsection{Plane-Based LiDAR--TCP Extrinsic Calibration and Chest Surface 3D Reconstruction}

For 3D reconstruction of the chest front surface and subsequent estimation of the initial probe pose, LiDAR points expressed in the LiDAR sensor frame $\{L\}$ must be mapped into the fixed robot base frame $\{B\}$. This requires the extrinsic transformation ${}^{C}\mathbf{T}_{L}$ between $\{L\}$ and the TCP frame $\{C\}$, since the robot kinematics provide ${}^{B}\mathbf{T}_{C}$. The overall mapping is thus

\begin{equation}
	\tilde{\mathbf{x}}_{B} 
	= {}^{B}\mathbf{T}_{C} \; {}^{C}\mathbf{T}_{L} \; \tilde{\mathbf{p}}_{L},
	\label{eq:lidar_to_base}
\end{equation}
\noindent where $\tilde{\mathbf{p}}_{L}$ is a homogeneous LiDAR point in $\{L\}$ and $\tilde{\mathbf{x}}_{B}$ its corresponding homogeneous coordinate in $\{B\}$.

We adopt a planar constraint strategy for this calibration in which a 600 mm × 900 mm flat board is placed within reach and the robot sweeps the LiDAR across it at multiple poses. The extrinsics are estimated by minimizing point to plane residuals, a common method for range sensor calibration that avoids external metrology and fiducials. Prior works~\cite{lembonoSCALARSimultaneousCalibration2018,lembonoSCALARSimultaneousCalibration2019,sharifzadehRobustHandeyeCalibration2020} employing this strategy were primarily developed for high-precision laser range finders, which exhibit low noise and stable beam geometry. Unlike multi-plane approaches or joint robot–sensor identification methods such as the SCALAR variants \cite{lembonoSCALARSimultaneousCalibration2018,lembonoSCALARSimultaneousCalibration2019}, we perform a separate extrinsics-only session with a single flat board. In this approach, plane fitting is applied with pose diversity and degeneracy checks to avoid near-parallel or collinear configurations. Finally, the transform is estimated directly to $\{C\}$ rather than to the flange. This reduces accumulated errors and ensures that the calibration is expressed directly in $\{C\}$ used for probe control, while retaining the simplicity and low setup cost of prior single plane methods \cite{sharifzadehRobustHandeyeCalibration2020}. In contrast to laser-based work, our method is adapted for low-cost 2D LiDARs, introducing angular filtering to mitigate scan non-uniformity, robustness measures for noisier point returns, and a lightweight single-board setup that simplifies calibration.

\begin{figure}[t!]
	\centering
	\includegraphics[width=0.48\textwidth, trim={0cm 0.0cm 0cm 0cm}, clip]{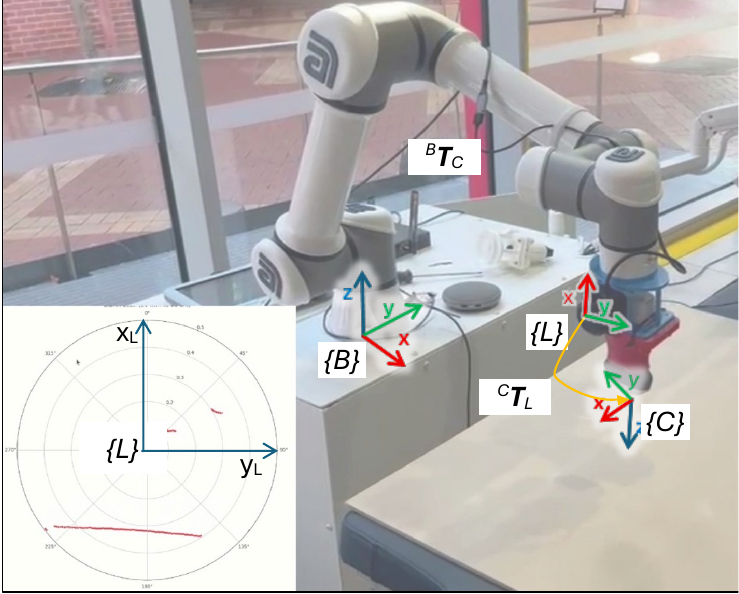}
	\captionsetup{width=\linewidth}
	\caption{Calibration setup showing the robot base $\{B\}$, TCP $\{C\}$, and LiDAR $\{L\}$ frames with the flat board as calibration plane. 
		LiDAR polar coordinates are defined with $0\degree$ along  $+\mathbf{x}_L$ and $90\degree$ along  $+\mathbf{y}_L$; only the sector $[135\degree,225\degree]$ is retained to capture board points for calibration.}
	\label{fig:fig2}
\end{figure}

The calibration setup is illustrated in~\autoref{fig:fig2}, with the LiDAR mounted rigidly to the probe holder and a flat board, which serves as the calibration target, positioned such that the LiDAR scanning points lie on the same plane. Robot kinematics provide the known transform ${}^{B}\mathbf{T}_{C}\in SE(3)$, and the goal of calibration is to estimate the extrinsic ${}^{C}\mathbf{T}_{L}\in SE(3)$. Following the convention in~\autoref{fig:fig2}, the LiDAR polar angle $\theta$ is defined such that $0\degree$ lies along the positive $\mathbf{x}_L$ axis and $90\degree$ along the positive $\mathbf{y}_L$ axis of frame $\{L\}$. 
A raw range return $(r,\theta)$ is converted as
\begin{equation}
	\mathbf{p}^{L}(r,\theta) =
	\begin{bmatrix}
		r\cos\theta\\
		r\sin\theta\\
		0
	\end{bmatrix},\qquad
	\tilde{\mathbf{p}}^{L}=\begin{bmatrix}\mathbf{p}^{L}\\1\end{bmatrix},
	\label{eq:polar}
\end{equation}
where $\mathbf{p}^{L}\in\mathbb{R}^3$ is the LiDAR point in frame $\{L\}$ and $\tilde{\mathbf{p}}^{L}\in\mathbb{R}^4$ its homogeneous form. During calibration, only the sector $\theta\in[135\degree,225\degree]$ is retained to restrict measurements to the calibration board.  

The robot is moved to $K=20$ distinct poses $\{{}^{B}\mathbf{T}_{C}^{(k)}\}_{k=1}^{K}$, providing sufficient positional and orientational diversity to fully determine the extrinsics while avoiding unnecessary data collection overhead. At each robot pose, multiple LiDAR scans are collected within the selected sector. The raw polar measurements are converted to Cartesian points $\mathbf{p}^{L}$ using~\autoref{eq:polar}, after which a RANSAC-based 2D line fitting procedure~\cite{fischlerRandomSampleConsensus1981} is applied. This extracts the dominant line corresponding to the intersection of the LiDAR scan plane ($z_L{=}0$, since the sensor acquires points only in its local $x_L$-$y_L$ plane) with the calibration board, while rejecting noisier point returns. The inlier set of LiDAR points at pose $k$ after line extraction is denoted $\{\mathbf{p}^{L}_{i}\}_{i\in\mathcal{I}_k}$, where $i$ indexes individual LiDAR points within the scan at pose $k$ and $\mathcal{I}_k$ is the set of indices corresponding to LiDAR points classified as inliers after angular filtering and line fitting.

Let the calibration plane in $\{B\}$ be
\begin{equation}
	\mathbf{n}_B^{\top}\mathbf{x}_B + d_B = 0, \qquad \|\mathbf{n}_B\|=1,
	\label{eq:plane}
\end{equation}
where $\mathbf{n}_B\in\mathbb{R}^3$ is the unit normal and $d_B\in\mathbb{R}$ the offset. Each LiDAR point is mapped into $\{B\}$ as
\begin{equation}
	\tilde{\mathbf{x}}_{B,i}^{(k)} = {}^{B}\mathbf{T}_{C}^{(k)}\; {}^{C}\mathbf{T}_{L}\; \tilde{\mathbf{p}}^{L}_{i},\qquad
	\mathbf{x}_{B,i}^{(k)}=\big[\tilde{\mathbf{x}}_{B,i}^{(k)}\big]_{1:3},
	\label{eq:transform}
\end{equation}
\noindent where $\tilde{\mathbf{x}}_{B,i}^{(k)}\in\mathbb{R}^4$ is the homogeneous coordinate and $\mathbf{x}_{B,i}^{(k)}\in\mathbb{R}^3$ the Cartesian point in $\{B\}$. The point-to-plane residual is
\begin{equation}
	r_{i}^{(k)} = \mathbf{n}_B^{\top}\mathbf{x}_{B,i}^{(k)} + d_B ,
	\label{eq:residual}
\end{equation}
which ideally equals zero for all $i\in\mathcal{I}_k$.  

We parameterize ${}^{C}\mathbf{T}_{L}$ by a rotation vector $\boldsymbol{\omega}\in\mathbb{R}^3$ and translation $\mathbf{t}\in\mathbb{R}^3$, and represent the plane normal as $\mathbf{n}_B=\mathbf{v}/\|\mathbf{v}\|$ for unconstrained $\mathbf{v}\in\mathbb{R}^3$. The unknown parameter vector is
\[
\boldsymbol{\xi} = \begin{bmatrix}\boldsymbol{\omega}^{\top} & \mathbf{t}^{\top} & \mathbf{v}^{\top} & d_B \end{bmatrix}^{\top}.
\]
The calibration is then formulated as a robust nonlinear least-squares problem
\begin{equation}
	\min_{\boldsymbol{\xi}}\;\sum_{k=1}^{K}\sum_{i\in\mathcal{I}_k}\rho\!\left(\mathbf{n}_B(\mathbf{v})^{\top}\mathbf{x}_{B,i}^{(k)}(\boldsymbol{\omega},\mathbf{t}) + d_B\right),
	\label{eq:opt}
\end{equation}
where $\rho(\cdot)$ is a robust loss (Cauchy). Although the LiDAR provides only planar measurements in its local frame ($z_L=0$), each point can be embedded into 3D by expressing it in homogeneous form $\tilde{\mathbf{p}}^{L}$ and transforming it sequentially through ${}^{C}\mathbf{T}_{L}$ and the known ${}^{B}\mathbf{T}_{C}^{(k)}$. This operation, shown in \autoref{eq:transform}, maps the 2D scan into the robot base frame $\{B\}$ for pose $k$. When repeated over multiple diverse robot poses, the LiDAR\rq s scanning plane is effectively swept through 3D space, generating a family of lines that all lie on the calibration board. These intersections provide sufficient geometric constraints to estimate both the extrinsic transform ${}^{C}\mathbf{T}_{L}$ and the board parameters $(\mathbf{n}_B,d_B)$ despite the LiDAR being only 2D.

To assess calibration quality, we report both per--pose and overall residual statistics. For each robot pose $k$, the RMS of the point-to-plane residuals $r_i^{(k)}$ in \autoref{eq:residual} quantifies local fit accuracy. Aggregating across all poses yields the overall RMS residual, which reflects the global consistency of ${}^{C}\mathbf{T}_{L}$. Calibration stability is further characterized by summary statistics across poses, including the mean, extrema, and variability of the per--pose RMS values. In addition, parameter covariance is approximated via the Gauss--Newton Jacobian at the solution of \autoref{eq:opt}, yielding one--sigma uncertainties for the estimated extrinsic translation and rotation parameters. Together, these metrics quantify the accuracy and robustness of the LiDAR--TCP calibration.

\begin{figure}[t!]
	\centering
	\includegraphics[width=0.485\textwidth, trim={0cm 0.0cm 0cm 0cm}, clip]{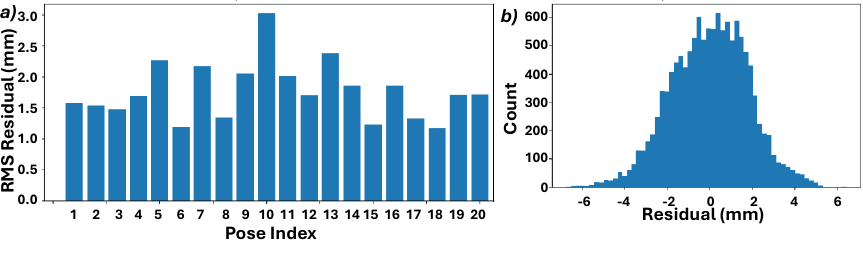}
	\captionsetup{width=\linewidth}
	\caption{Calibration accuracy results. a) Per-pose RMS residuals across 20 robot poses, showing consistent $<$ 3 mm errors. b) Histogram of all point-to-plane residuals, tightly centered around zero with most inliers within $\pm$ 3 mm.}
	\label{fig:fig4}
\end{figure}

As shown in \autoref{fig:fig4}a, the calibration achieved a mean per-pose RMS of $1.77~$mm (range $1.17$~mm -- $3.03~$mm), with an overall RMS residual of $1.82~$mm. The residual distribution in \autoref{fig:fig4}b was tightly centered near zero, with most inliers lying within $\pm 3~$mm, indicating accuracy without systematic bias. Uncertainty analysis further confirmed that the estimated extrinsic transform had one-sigma translation uncertainties within $1.1~$mm and rotation uncertainties within $0.2\degree$, demonstrating sufficient accuracy and stability for downstream probe initialization.

\begin{figure*}[t!]
	\centering
	\includegraphics[width=0.95\textwidth, trim={0.0cm 0.0cm 0.0cm 0.0cm}, clip]{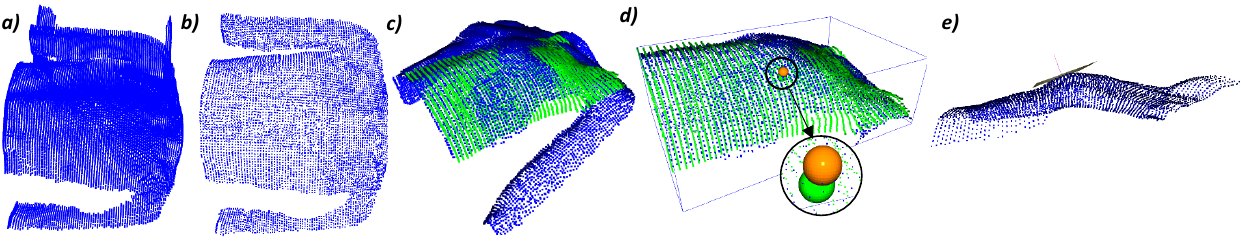}
	\captionsetup{width=\linewidth}
	\caption{Point clouds from mannequin at key stages: (a) raw 3D point cloud; (b) after preprocessing; (c) alignment of template (green) with preprocessed (blue) cloud; (d) cropped chest surface with template point (green sphere) and corresponding point (orange sphere); (e) estimated initial probe point with local PCA plane and surface normal vector.}
	\label{fig:fig_6}
\end{figure*}

With ${}^{C}\mathbf{T}_{L}$ estimated, we reconstruct the chest\rq s front surface by executing two non-contact linear sweeps with the robot. In our setup, the bed\rq s longitudinal axis is approximately parallel to the robot-base $y$-axis ($y_B$). Accordingly, the end-effector traverses two paths approximately parallel to $y_B$ over the left chest's front surface, each with a slightly different orientation to better capture the left-heart region, while maintaining a safe clearance $d_{\mathrm{safe}}$ from the body. During motion, the LiDAR continuously reports polar measurements $(r,\theta)$ within the sector $[135\degree,225\degree]$, which are converted to Cartesian points $\mathbf{p}^{L}(r,\theta)$ through \autoref{eq:polar} and mapped into the base frame using the time-stamped kinematic transform ${}^{B}\mathbf{T}_{C}(t)$ and the calibrated extrinsics ${}^{C}\mathbf{T}_{L}$ as in \autoref{eq:transform}. Accumulating samples over the two passes yields an unfiltered point cloud
\[
\mathcal{P}_{\mathrm{raw}}=\bigcup_{k}\bigcup_{i\in\mathcal{I}_k}\mathbf{x}_{B,i}^{(k)}\subset\mathbb{R}^3,
\]
which forms a sparse 3D reconstruction that includes the chest's front surface as well as incidental structures such as the arms and the bed, as shown in~\autoref{fig:fig_6}a.

\begin{figure}[b!]
	\centering
	\includegraphics[width=0.485\textwidth, trim={0.55cm 0.3cm 0.32cm 0.4cm}, clip]{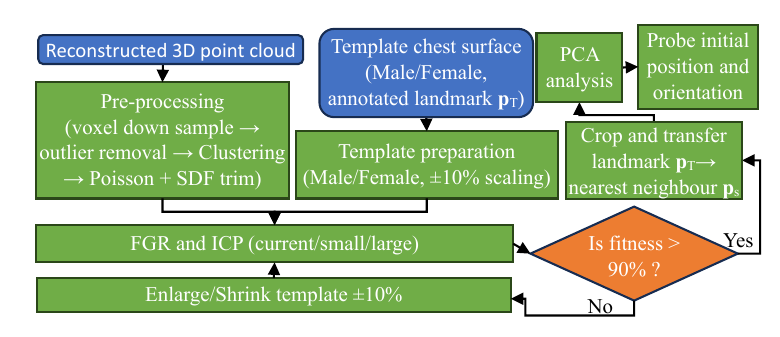}
	\captionsetup{width=\linewidth}
	\caption{Framework for initial probe pose estimation via non-rigid template matching with enlarge/shrink alignment loop and PCA-based orientation estimation.}
	\label{fig:fig_5}
\end{figure}

\subsection{Non-rigid Template Matching Framework for Initial Probe Pose Estimation}
Following 3D reconstruction, we apply the non-rigid template matching framework illustrated in~\autoref{fig:fig_5} to automatically estimate the probe's initial pose. Template matching has been applied in both general object localization~\cite{lopezSpecificObjectFinding2023} and robotic ultrasound systems with depth camera-based sensing~\cite{roshanSensorFusionApproach2025} and we adapt this in the proposed approach. As shown in~\autoref{fig:fig_6}b, the reconstructed 3D point cloud is first preprocessed using voxel downsampling to ensure uniform point density, followed by statistical and radius outlier removal, density-based clustering, and Poisson surface reconstruction with signed-distance trimming. These steps remove floating artifacts and non-body structures while preserving chest geometry. Predefined chest surface templates in the form of point clouds (male and female variants), each annotated with an ideal initial probe point $\mathbf{p}_T \in \mathbb{R}^3$, were employed as references. To accommodate inter-subject variability, each template was scaled about its centroid to generate three variants: the original size, +10\%, and -10\%. For each variant, surface normals and fast point feature histograms (FPFH) were computed, and fast global registration (FGR) followed by iterative closest point (ICP) refinement was applied to obtain correspondences. If any variant achieved an alignment fitness $\geq$90\%, that result was selected, resulting in the aligned template and preprocessed point cloud as shown in~\autoref{fig:fig_6}c. Otherwise, the variant with the highest fitness was used to guide further scaling: either enlarged or shrunken in additional 10\% increments. This process was repeated iteratively until the target fitness threshold of 90\% was reached. 

After alignment, the template point $\mathbf{p}_T$ was mapped to its corresponding point $\mathbf{p}_S$ on the preprocessed point cloud, identified as the nearest neighbor to $\mathbf{p}_T$, yielding the initial probe position, as shown in~\autoref{fig:fig_6}d. Maintaining the probe normal to the chest surface offers an anatomically neutral and clinically ideal~\cite{mitchellGuidelinesPerformingComprehensive2019} starting orientation for subsequent teleoperated adjustments. To derive that orientation, principal component analysis (PCA) was applied to the $k=30$ nearest neighbors of $\mathbf{p}_S$, yielding the eigenvector associated with the smallest eigenvalue as the local surface normal $\mathbf{n}_S$, as shown in~\autoref{fig:fig_6}e. The initialization pose for the probe was thus defined for automated positioning. An open-source implementation of the complete approach, including LiDAR calibration, reconstruction, and non-rigid template matching, is available at~\textit{[GitHub link BLINDED FOR REVIEW]}.

\section{Evaluations}
\subsection{Mannequin-Based Geometric Validation of LiDAR-based 3D Reconstruction}

\begin{figure}[b!]
	\centering
	\includegraphics[width=0.5\textwidth, trim={0.0cm 0.0cm 0.0cm 0.0cm}, clip]{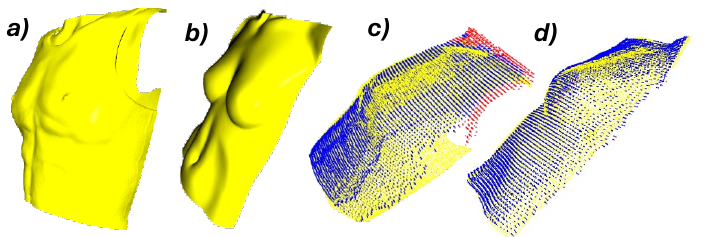}
	\captionsetup{width=\linewidth}
	\caption{Cropped chest surfaces of the male (a) and female (b) mannequins. In (c) and (d), the LiDAR point clouds for the male and female mannequins are color-coded by distance to the GT surface (GT shown in yellow): blue $<2$~mm, green $[2,8]$~mm, red $>8$~mm.}
		\label{fig:fig_7}
\end{figure}

The purpose of this experiment was to quantify the geometric accuracy of the LiDAR-based reconstruction against high-resolution ground truth (GT) point clouds. The front chest surfaces of male and female mannequins were manually scanned using a high-resolution Revopoint 3D scanner, yielding the GT point cloud $\mathcal{P}_{\mathrm{GT}}$ for each mannequin, as shown in~\autoref{fig:fig_7}a-b. For each mannequin, 3D reconstructions were obtained using five distinct LiDAR configurations that varied both the sensor's distance from the chest surface and its orientation to assess robustness, resulting in ten LiDAR-based point clouds, each denoted as $\mathcal{P}_{\mathrm{L}}$. These point clouds were manually cropped in MeshLab to remove background and non-chest structures.  

The cleaned $\mathcal{P}_{\mathrm{L}}$ was then aligned to $\mathcal{P}_{\mathrm{GT}}$ using FGR followed by local refinement with ICP. To verify alignment quality prior to geometric error analysis, we computed the final ICP fitness score $f_{\mathrm{ICP}}$ and the inlier root-mean-square error $e_{\mathrm{ICP}}$. In addition, we calculated the percentage of LiDAR points within an 8 mm tolerance of the GT point cloud, denoted as $c_{\mathrm{L}}$. This threshold corresponds to the native spatial resolution of the RPLIDAR C1 ($0.72\degree$ angular step, projecting to 6-10 mm lateral spacing at the tested stand-off distances) and represents the smallest reliable discrimination of the LiDAR. For reconstruction accuracy, we calculated the RMSE of nearest-neighbor distances from $\mathcal{P}_{\mathrm{L}}$ to $\mathcal{P}_{\mathrm{GT}}$, denoted $e_{\mathrm{RMSE}}$, along with the 95th percentile error $e_{95}$. Results were aggregated across repeats to compute the mean and standard deviation per condition, and overall statistics across all ten trials were reported.

\subsection{Initial Probe Position Validation on Human Subjects}
The objective of this evaluation was to quantify how accurately the complete end-to-end LiDAR-based reconstruction and non-rigid template matching approach identifies an initial probe position.

\begin{figure}[t!]
	\centering
	\includegraphics[width=0.5\textwidth, trim={0.0cm 0.0cm 0.0cm 0.0cm}, clip]{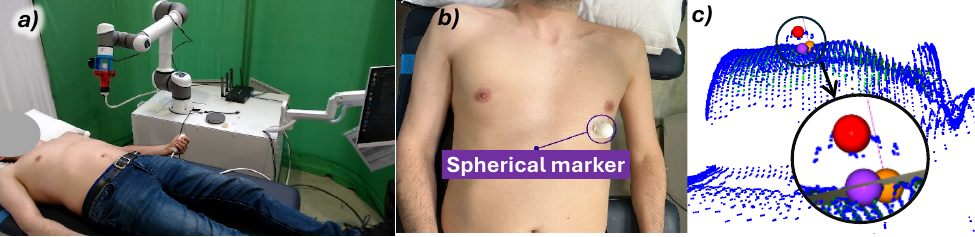}
	\captionsetup{width=\linewidth}
	\caption{Human evaluation setup and outcomes. (a) Volunteer positioned supine during LiDAR-based chest surface reconstruction. (b) 3D-printed spherical marker placed at the point of maximal impulse (PMI) as ground-truth reference. (c) Aligned raw (blue) and preprocessed  (green) point clouds with the selected point on the marker (red), the approach estimated initial probe position (orange), and the projected intersection point on the local chest plane (purple).}
	\label{fig:fig_8}
\end{figure}

We recruited $N{=}5$ healthy volunteers under~\textit{[BLINDED FOR REVIEW]} institutional ethics approval (protocol ID~\textit{[BLINDED FOR REVIEW]}). As shown in~\autoref{fig:fig_8}a, each volunteer was positioned supine. A 3D-printed spherical marker of 30~mm diameter was used as a physical ground-truth target. The size was chosen to ensure robust detectability in the LiDAR point cloud while remaining small enough to be comfortably placed on the chest surface without obstructing the scanning region. The marker was positioned at the point of maximal impulse (PMI), corresponding to the cardiac apex beat typically located on the left chest between the fifth and sixth ribs~\cite{williamsPhysicalExamPresence2023}, approximately in line with the nipple, as shown in~\autoref{fig:fig_8}b. This site represents the strongest outward cardiac motion and provides a clinically practical starting point for echocardiography. 

For each volunteer, two linear LiDAR sweeps were performed for 3D reconstruction, and the procedure was repeated across three independent trials. In each trial, before pre-processing and applying the non-rigid template matching framework, a point on the top surface of the spherical marker was manually selected on the raw LiDAR point cloud via a single-point pick, thereby recording the marker location $\mathbf{p}_M^{\text{sphere}} \in \mathbb{R}^3$ in the robot base frame $\{B\}$. This selection was made on the raw point cloud to ensure the marker was preserved, since subsequent preprocessing could otherwise remove it. The complete approach described in Section~II was then applied to estimate the initial probe position and orientation for comparison against the ground-truth marker.

To validate the accuracy of the estimated initial point $\mathbf{p}_S$ produced by the proposed approach, the comparison was made against the physical ground-truth marker location. However, since the spherical marker protrudes above the chest surface, direct comparison could introduce bias. As shown in~\autoref{fig:fig_8}c, to address this bias, $\mathbf{p}_M^{\text{sphere}}$ was orthogonally projected onto the local chest surface at $\mathbf{p}_S$ as

\begin{equation}
	\mathbf{p}_M^{\parallel} = \mathbf{p}_S + \left( \mathbf{I} - \hat{\mathbf{n}}_S \hat{\mathbf{n}}_S^\top \right) (\mathbf{p}_M^{\text{sphere}} - \mathbf{p}_S),
\end{equation}

\noindent where $\hat{\mathbf{n}}_S$ is the local surface normal at $\mathbf{p}_S$. This yielded a corrected, skin-level ground-truth location $\mathbf{p}_M^{\parallel} \in \mathbb{R}^3$. Thereafter, the tangential positional error was calculated as
\begin{equation}
	e_{\parallel} \;=\; \left\lVert \mathbf{p}_M^{\parallel} - \mathbf{p}_S \right\rVert_2 \;\; .
	\label{eq:eparallel}                                                                                                                                                                                                                                                                      
\end{equation}

Repeatability was assessed in two ways: first, by computing the within-subject standard deviation of $e_{\parallel}$ across the three repeated trials for each participant, and second, by calculating the intraclass correlation coefficient (ICC) over the full $5\times3$ set of $e_{\parallel}$ values. In addition, alignment quality was characterized using the final ICP fitness $f_{\mathrm{ICP}}$, together with the inlier root-mean-square error $e_{\mathrm{ICP}}$.

\begin{table}[t!]
	\centering
	\caption{Quantitative results of mannequin evaluation (mean $\pm$ SD across five trials per mannequin).}
	\label{tab:mannequin_results}
	\begin{tabularx}{\linewidth}{l *{5}{>{\centering\arraybackslash}X}}
		\hline
		Condition & $f_{\mathrm{ICP}}$ & $e_{\mathrm{ICP}}$ [mm] & $c_{\mathrm{L}}$ [\%] & $e_{\mathrm{RMSE}}$ [mm] & $e_{95}$ [mm] \\
		\hline
		Male (N=5)   & $0.974 \pm 0.012$ & $3.01 \pm 0.10$ & $95.1 \pm 1.3$ & $2.97 \pm 0.08$ & $5.18 \pm 0.12$ \\
		Female (N=5) & $0.978 \pm 0.013$ & $2.62 \pm 0.14$ & $98.6 \pm 1.9$ & $2.59 \pm 0.02$ & $4.53 \pm 0.04$ \\
		Overall (N=10) & $0.976 \pm 0.012$ & $2.82 \pm 0.24$ & $96.8 \pm 2.4$ & $2.78 \pm 0.21$ & $4.86 \pm 0.36$ \\
		\hline
	\end{tabularx}
\end{table}

\begin{table}[b!]
	\centering
	\caption{Tangential error $e_{\parallel}$ (mm) for each participant, reported as mean $\pm$ standard deviation over three independent trials.}
	\label{tab:human_eval_results}
	\begin{tabularx}{\linewidth}{l *{5}{>{\centering\arraybackslash}X}}
		\hline
		& \textbf{P1} & \textbf{P2} & \textbf{P3} & \textbf{P4} & \textbf{P5} \\
		\hline
		Mean $\pm$ SD [mm] & $26.32 \pm 3.93$ & $24.48 \pm 2.81$ & $28.00 \pm 0.30$ & $23.75 \pm 3.15$ & $26.14 \pm 1.72$ \\
		\hline
	\end{tabularx}
\end{table}
\section{Results and Discussion}
In the mannequin evaluation, quantitative results from ten trials are summarized in \autoref{tab:mannequin_results}, separated by male and female mannequins with overall statistics reported for the combined dataset. Alignment quality was consistently high, with $f_{\mathrm{ICP}}$ exceeding 0.95 in all cases and $e_{\mathrm{ICP}}$ remaining below 3.1~mm. The coverage $c_{\mathrm{L}}$ confirmed that over 93\% of LiDAR points were within the 8~mm tolerance of the ground truth, ensuring robust alignment prior to error analysis. In terms of reconstruction accuracy, $e_{\mathrm{RMSE}}$ averaged 2.97~mm for male and 2.59~mm for female mannequins, with $e_{95}$ of 5.18~mm and 4.53~mm, respectively. The slightly lower errors for female scans likely reflect smoother chest geometry, which yields fewer local discontinuities during registration. In contrast, the male mannequin exhibits more pronounced anatomical contouring (sternum and rib cage), where minor mismatches in sampling and alignment accumulate as larger local deviations. Across all conditions, the overall reconstruction error remained low, with $e_{\mathrm{RMSE}}=2.78\pm0.21$~mm and $e_{95}=4.86\pm0.36$~mm. These trends are also visible in the color-coded error maps for one pair of alignment in~\autoref{fig:fig_7}c-d, where LiDAR point clouds for the male and female mannequins are shown relative to the GT surface: blue $<2$~mm, green $[2,8]$~mm, and red $>8$~mm. All these results confirm that the proposed method provides reliable and repeatable surface 3D reconstructions suitable for subsequent template-based initial probe pose estimation.

\begin{figure}[t!]
	\centering
	\includegraphics[width=0.5\textwidth, trim={0.0cm 0.0cm 0.0cm 0.0cm}, clip]{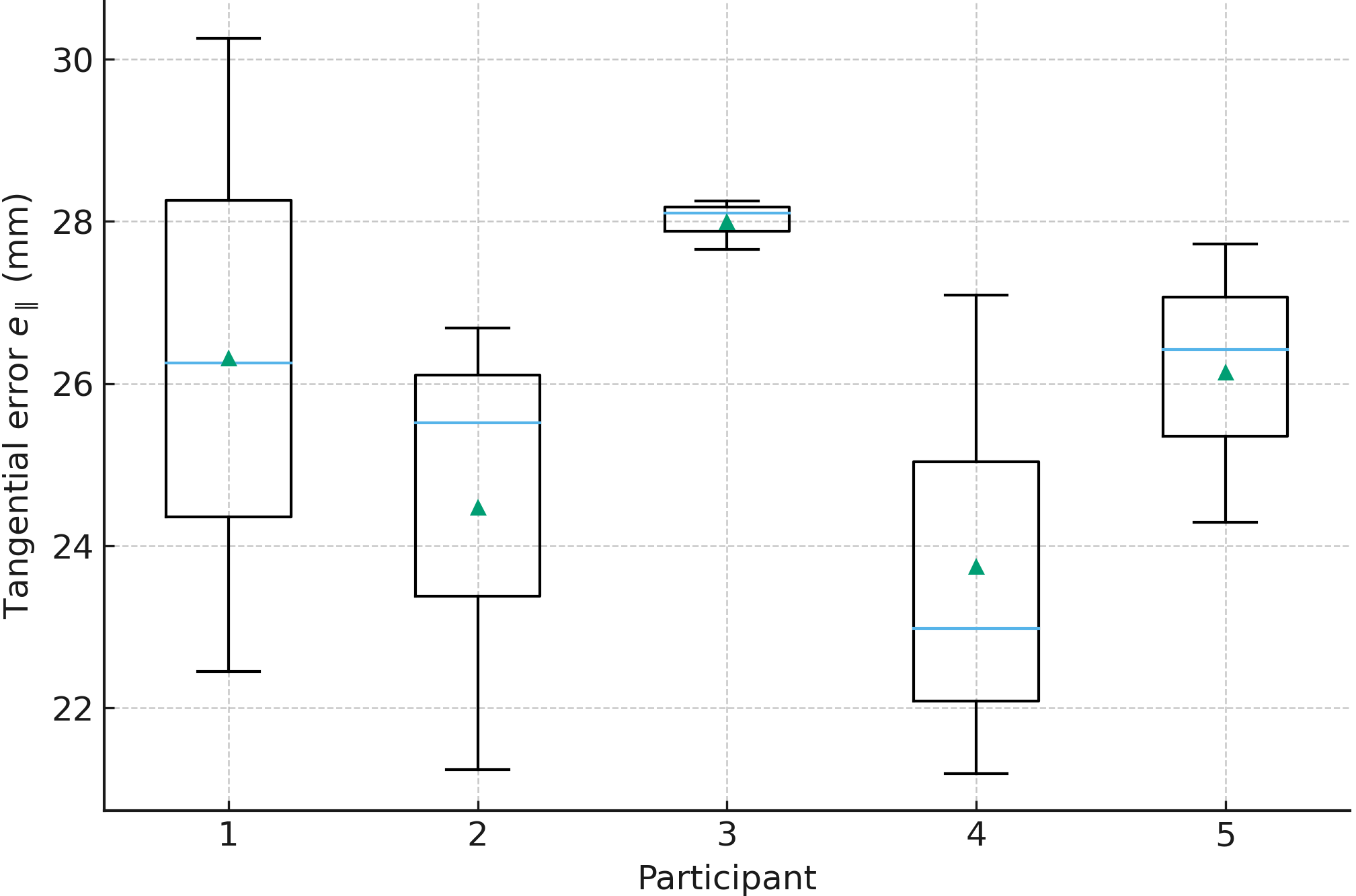}
	\captionsetup{width=\linewidth}
	\caption{Distribution of tangential error $e_{\parallel}$ across three trials per participant (P1--P5). The narrow spread within participants indicates good repeatability of the proposed initialization method.}
	\label{fig:fig_9}
\end{figure}

In the human subject evaluation, the $e_{\parallel}$ across all participants and trials ranged from 21--30~mm, with mean values clustering in the mid--20~mm range. Coarse placement errors are operationally acceptable, since routine echocardiography relies on sliding and minor probe adjustments to refine the acoustic window and achieve usable images~\cite{mitchellGuidelinesPerformingComprehensive2019}. Within-subject variability was small, with standard deviations between 0.3--3.9~mm as summarized in~\autoref{tab:human_eval_results}, and trial-level errors were tightly clustered around subject-specific means with minimal outliers as shown in~\autoref{fig:fig_9}. Alignment quality metrics further supported this stability, with $f_{\mathrm{ICP}}$ converging to 1.0 in all cases and $e_{\mathrm{ICP}}$ between 4--6~mm, confirming that the template registration was geometrically well constrained. Taken together, these results demonstrate that the system reliably converges to the same initial probe position on repeated runs, and that the absolute offsets of 21--30~mm are most likely attributable to inter-subject anatomical variability rather than systematic errors in the proposed algorithm.

Overall, the findings indicate that the LiDAR-based method provides a consistent and robust initial probe pose estimation, with initialization errors well within the anatomical variability encountered in clinical echocardiography.

\section{Conclusion and Future Work}

This work presented the first demonstration of using a robot-mounted 2D LiDAR to automate probe pose initialization in teleoperated echocardiography. A plane-based calibration strategy enabled extrinsic accuracy of less than $2$ mm in translation and under $0.2\degree$ rotation, after which two linear sweeps for LiDAR reconstruction produced consistent 3D models of the chest front surface. A non-rigid template-matching framework with sex-specific chest front surface priors and PCA-based orientation estimation then localized a clinically meaningful initial probe pose for placement. Quantitative validation on mannequins confirmed reconstruction accuracy within 3~mm RMSE and $<5$~mm at the 95th percentile, while human trials demonstrated repeatable convergence to subject-specific initial probe position with lateral deviations of 20--30~mm, consistent with the anatomical variability of the cardiac acoustic window. Together, these findings demonstrate that LiDAR-based initial probe pose estimation offers a robust strategy for automating probe placement in tele-echocardiography.

Future work will extend this foundation in three directions. First, LiDAR scanning trajectories will be optimized adaptively to patient size, with the dual goals of reducing acquisition time and limiting irrelevant point captures or noise. Second, we will define and evaluate strategies for safe and efficient motion from a rest position to the identified probe initial pose, ensuring clearance of the human and bed while maintaining clinical safety margins. Finally, we plan a full clinical study to compare time-to-initialization between the proposed LiDAR-based automation and manual sonographer manipulation, as well as to capture structured feedback from sonographers regarding usability and integration into routine teleoperated workflows. These efforts will advance the system from proof-of-concept toward clinical translation, ultimately supporting more efficient and accessible tele-echocardiography.

\bibliographystyle{IEEEtran}
\bibliography{IEEE}

\end{document}